\documentclass[10pt,twocolumn,letterpaper]{article}

\usepackage[pagenumbers]{cvpr} % To force page numbers, e.g. for an arXiv version

\usepackage{graphicx}
\usepackage{amsmath}
\usepackage{amssymb}
\usepackage{booktabs}

\usepackage{paralist, tabularx}
\usepackage{enumitem}

\usepackage[pagebackref,breaklinks,colorlinks,citecolor=blue,linkcolor=blue,urlcolor=blue]{hyperref}

\usepackage[capitalize]{cleveref}
\crefname{section}{Sec.}{Secs.}
\Crefname{section}{Section}{Sections}
\Crefname{table}{Table}{Tables}
\crefname{table}{Tab.}{Tabs.}

\newcommand\blfootnote[1]{%
  \begingroup
  \renewcommand\thefootnote{}\footnote{#1}%
  \addtocounter{footnote}{-1}%
  \endgroup
}

\def\cvprPaperID{564} % *** Enter the CVPR Paper ID here
\def\confName{CVPR}
\def\confYear{2023}

\begin{document}

\title{Hi-LASSIE: High-Fidelity Articulated Shape and Skeleton Discovery\\ from Sparse Image Ensemble}

\author{
Chun-Han Yao$^1$\textsuperscript{*} \hspace{3mm}
Wei-Chih Hung$^2$ \hspace{3mm}
Yuanzhen Li$^3$ \hspace{3mm}
Michael Rubinstein$^3$ \\
Ming-Hsuan Yang$^{134}$ \hspace{3mm}
Varun Jampani$^3$ \\\\
$^1$UC Merced \hspace{3mm}
$^2$Waymo \hspace{3mm} 
$^3$Google Research \hspace{3mm} 
$^4$Yonsei University
}

\twocolumn[{
\renewcommand\twocolumn[1][]{#1}
\maketitle
\begin{center}
\vspace{-3mm}
\centering
\includegraphics[width=.9\linewidth]{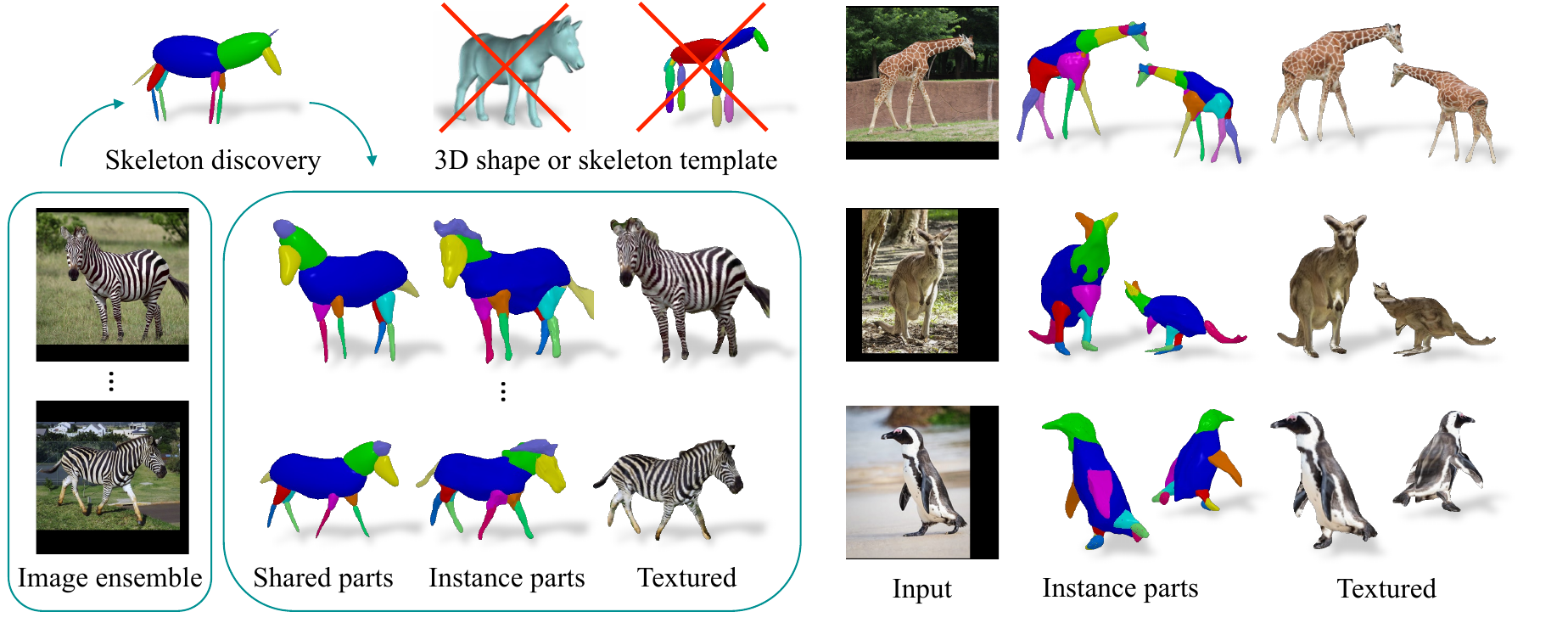}
\vspace{-3mm}
\captionof{figure}{\small \textbf{Hi-LASSIE overview and sample reconstructions.} Given 20-30 images of an articulated animal class, we first discover a generic 3D skeleton, then jointly optimize the camera viewpoints, skeleton articulations, as well as shared and instance-specific neural part shapes. Hi-LASSIE is able to produce high-fidelity shapes and texture without any pre-defined shape model or 3D skeleton annotations. The part-based representation also allows applications like animation and motion re-targeting.}
\label{fig:cover}
\end{center}
}]

\blfootnote{\textsuperscript{*}{Work done as a student researcher at Google.}}

%%%%%%%%% ABSTRACT
\begin{abstract}
\vspace{-3mm}
Automatically estimating 3D skeleton, shape, camera viewpoints, and part articulation from sparse in-the-wild image ensembles is a severely under-constrained and challenging problem. Most prior methods rely on large-scale image datasets, dense temporal correspondence, or human annotations like camera pose, 2D keypoints, and shape templates. We propose Hi-LASSIE, which performs 3D articulated reconstruction from only 20-30 online images in the wild without any user-defined shape or skeleton templates. We follow the recent work of LASSIE that tackles a similar problem setting and make two significant advances. First, instead of relying on a manually annotated 3D skeleton, we automatically estimate a class-specific skeleton from the selected reference image. Second, we improve the shape reconstructions with novel instance-specific optimization strategies that allow reconstructions to faithful fit on each instance while preserving the class-specific priors learned across all images. Experiments on in-the-wild image ensembles show that Hi-LASSIE obtains higher fidelity state-of-the-art 3D reconstructions despite requiring minimum user input. Project page: \url{chhankyao.github.io/hi-lassie/}

\vspace{-3mm}
\end{abstract}

%%%%%%%%%%%%%%%%%%%%%%%%%%%%%%%%%%%%%%%%%%%%%%%%%%%%%%%%%%%%
\vspace{-2mm}
\section{Introduction}
\vspace{-2mm}

\begin{figure*}[t!]
\centering
\includegraphics[width=.9\linewidth]{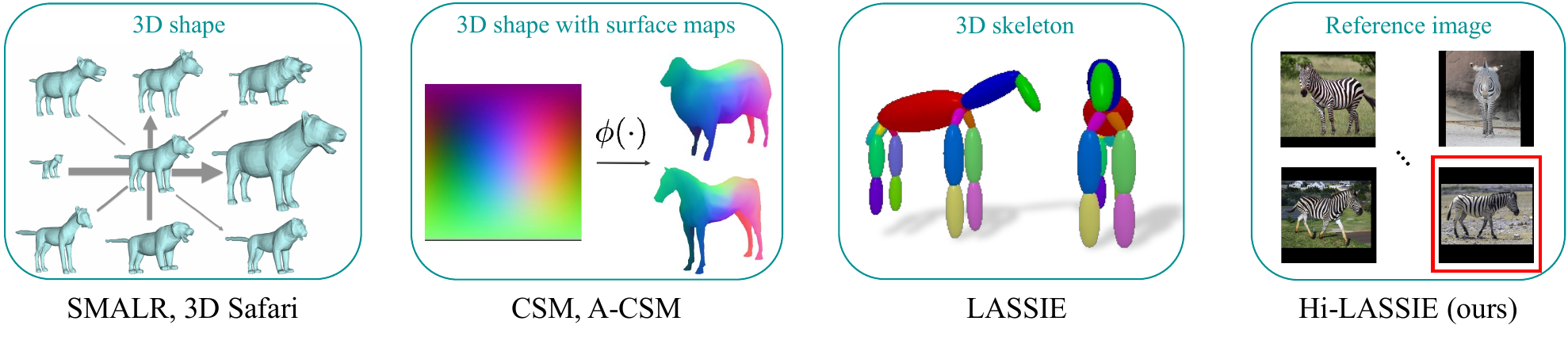}
\vspace{-3mm}
\caption{\textbf{User inputs across different techniques for articulated animal reconstruction.} Contrary to prior methods that leverage detailed 3D shapes or skeletons, Hi-LASSIE only requires the user to select a reference image where most animal body parts are visible.}
\vspace{-2mm}
\label{fig:user_templates}
\end{figure*}

%%%%%%%%%%%%%%%%%%%%%%%%%%%%%%%%%%%%%%%%%%%%%%%%%%%%%%%%%%%%
%%% Problem motivation and definition.
%
3D assets of articulated animals enable numerous applications in games, movies, AR/VR, etc.
However, building high-fidelity 3D models of articulated shapes like animal bodies is labor intensive either via manual creation or 3D scanning.
Recent advances in deep learning and 3D representations have significantly improved the quality of 3D reconstruction from images.
Much of the success depends on the availability of either rich 3D annotations or multi-view captures, both of which are not always available in a real-world scenario.
A more practical and scalable alternative is automatic 3D reconstruction from online images as it is straightforward to obtain image ensembles of any animal category (\eg, image search results).
In this work, we tackle a practical problem setting introduced in a recent work, LASSIE~\cite{yao2022lassie}, where the aim is to automatically estimate articulated 3D shapes with only a few (20-30) in-the-wild images of an animal species, without any image-level 2D or 3D annotations.

%%%%%%%%%%%%%%%%%%%%%%%%%%%%%%%%%%%%%%%%%%%%%%%%%%%%%%%%%%%%
%%% Challenges and briefly about issues in existing works.
%
This problem is highly under-constrained and challenging due to a multitude of variations within image ensembles.
% camera viewpoints, part articulations, textures, backgrounds, illuminations etc.
% Estimating articulated animal bodies from sparse image ensemble is a highly ambiguous and challenging task.
% %
In-the-wild images usually have diverse backgrounds, lighting, and camera viewpoints.
Moreover, different animal instances can have distinct 2D appearances due to pose articulations, shape variations, and surface texture variations (skin colors, patterns, and lighting).
As shown in Fig.~\ref{fig:user_templates}, early approaches in this space try to simplify the problem with some user-defined 3D templates, hurting the generalization of those techniques to classes where such templates are not always readily available.
In addition, most of these methods (except LASSIE~\cite{yao2022lassie}) assume either large-scale training images of each animal species~\cite{kulkarni2019canonical,kulkarni2020articulation,zuffi2019three} or per-image 2D keypoint annotations~\cite{zuffi2018lions}, which also limits their scalability.

%%%%%%%%%%%%%%%%%%%%%%%%%%%%%%%%%%%%%%%%%%%%%%%%%%%%%%%%%%%%
%%% Introduction to our technique
%
In this work, we propose a more practical technique that does not require any 3D shape or skeleton templates.
% In this work, we present Hi-LASSIE to produce \textbf{Hi}gh-qua\textbf{L}ity \textbf{A}rticulated \textbf{S}hapes from \textbf{S}parse \textbf{I}mage \textbf{E}nsemble.
%
Instead, as illustrated in Fig.~\ref{fig:user_templates}, the user simply has to select a reference image in the ensemble where all the animal parts are visible.
We achieve this by providing two key technical advances over LASSIE: 1) 3D skeleton discovery and 2) instance-specific optimization strategies.
Our overall framework, named Hi-LASSIE, can produce \textbf{Hi}gher-fidelity articulated shapes than \textbf{LASSIE}~\cite{yao2022lassie} while requiring minimal human input.
%
% To make our method wildly applicable to more diverse object classes and flexible to fit individual instances, we further remove the assumptions in LASSIE: 1) a generic 3D skeleton is given and 2) the 3D part shapes are shared by all instances.
% %
% That is, our framework does not require any 3D skeleton, shape template, or image-level annotations.
%
% %
% Similar to LASSIE, we adopt a part-based representation build upon a 3D skeleton, which enables flexible articulations under pose and shape constraints.
% %
% Nonetheless, we aim to discover 3D skeleton automatically from an image ensemble, making it even more challenging.
%
%%%%%%%%%%%%%%%%%%%%%%%%%
%%% Insights for 3D skeleton discovery.
Our key insight is to exploit the 2D part-level correspondences for 3D skeleton discovery.
Recent works~\cite{amir2021deep,tumanyan2022splicing} observe that the deep features extracted from a self-supervised vision transformer (ViT)~\cite{dosovitskiy2020image} like DINO-ViT~\cite{caron2021emerging} can provide good co-part segmentation across images.
We further exploit such features to reason about part visibility and their symmetry.
At a high level, we first obtain a 2D skeleton using the animal silhouette and part clusters~\cite{amir2021deep} obtained from DINO features~\cite{caron2021emerging}.
We then uplift this 2D skeleton into 3D using symmetric part information that is present in the deep DINO features.
Fig.~\ref{fig:cover} shows the skeleton for zebra images discovered by Hi-LASSIE.
%
% %
% Recent methods~\cite{amir2021deep,tumanyan2022splicing} demonstrate that the self-supervisory deep features from a vision transformer~\cite{dosovitskiy2020image} (DINO-ViT~\cite{caron2021emerging}) provide dense 2D correspondence between images of similar animal classes.
% %
% Our key insight is that such features can not only be used to obtain foreground/part masks, but also serve as visibility and symmetry information which are crucial to 3D skeleton extraction and global pose estimation.
% %
% As shown in Fig.~\ref{fig:cover} (left), given a reference image where most body parts are visible (non-occluded), we first extract 2D skeleton points from the silhouette, then initialize the 3D joint coordinates and simple part shapes based on distance transform and DINO features.
%
%%%%%%%%%%%%%%%%%%%%%%%%%
%%% On better optimization strategies
%
%
Similar to LASSIE~\cite{yao2022lassie}, we leverage 3D part priors (learned from and shared across instances) and the discovered 3D skeleton to regularize the articulated shape learning.
%
% To reconstruct the 3D articulated shapes, we start with the discovered skeleton and then optimize the neural part surfaces shared by all instances as well as instance-specific parameters like camera viewpoints, bone transformations, and detailed part deformations. 
%
Furthermore, we design three novel modules to increase the quality of output shapes: 1) High-resolution optimization by zooming in on individual parts, 2) Surface feature MLPs to densely supervise the neural part surface learning, and 3) Frequency-based decomposition of part surfaces for shared and instance-specific components.
Note that Hi-LASSIE can generalize to diverse animal species easily as it does not require any image annotations or category-specific templates.

%%%%%%%%%%%%%%%%%%%%%%%%%%%%%%%%%%%%%%%%%%%%%%%%%%%%%%%%%%%%
%%%% Summary of experiments, datasets and results.
%
We conduct extensive experiments on the Pascal-Part~\cite{chen2014detect} and LASSIE~\cite{yao2022lassie} image ensembles, which contain in-the-wild images of various animal species like horse, elephant, and penguin.
Compared with LASSIE and other baselines, we achieve higher reconstruction accuracy in terms of keypoint transfer, part transfer, and 2D IOU metrics.
Qualitatively, Hi-LASSIE reconstructions show considerable improvement on 3D geometric and texture details as well as faithfulness to input images.
Finally, we demonstrate several applications like animation and motion re-targeting enabled by our 3D part representation.
Fig.~\ref{fig:cover} (right) shows some Hi-LASSIE 3D reconstructions for different animal species.
The main contributions of this work are:
%
%\begin{itemize}[leftmargin=*]
%\vspace{-2mm}
%
%MH: you can adjust this later if space allows
\setdefaultleftmargin{0em}{2em}{}{}{}{}
\begin{compactitem}
\item To our best knowledge, Hi-LASSIE is the first approach to discover 3D skeletons of articulated animal bodies from in-the-wild image ensembles without using any image-level annotations. We show that the discovered 3D skeleton can faithfully fit all instances in the same class and effectively regularize the 3D shape optimization.
\item Hi-LASSIE includes several novel optimization strategies that makes the output shapes richer in 3D details and more faithful to each image instance than prior methods.
% We propose an optimization framework, Hi-LASSIE, by using a 3D part representation built upon the discovered 3D skeleton. We decompose the 3D part surfaces in the frequency space, allowing instance-specific deformations while preserving the common base shapes shared across all instances. Additional optimization techniques like zoomed-in part rendering and surface features MLPs are also shown to improve the output shape details.
%
\item Extensive results on multiple animal classes and datasets demonstrate the state-of-the-art performance of Hi-LASSIE while requiring less user inputs than prior works.
\end{compactitem}
\vspace{-2mm}
\section{Related Work}
\vspace{-2mm}

%%%%%%%%%%%%%%%%%%%%%%%%%%%%%%%%%%%%%%%%%%%%%%%%%%%%%%%%%%%%
{\noindent \textbf{Animal pose and shape estimation.}~}
3D pose and shape estimation of animal bodies from in-the-wild images is quite challenging considering the diverse 2D appearance across different instances, viewpoints, and articulations, among others.
Most mesh reconstruction methods~\cite{kulkarni2019canonical,li2020self,goel2020shape,tulsiani2020implicit,ye2021shelf} mainly deal with objects with simple or rigid shapes (\eg, birds and cars), which cannot be applied to the highly-articulated animal bodies.
Recent articulation-aware approaches address a different (often simplified) scenario by leveraging a pre-defined statistical model~\cite{zuffi20173d,zuffi2018lions, zuffi2019three,rueegg2022barc}, 3D shape/skeleton templates~\cite{kulkarni2020articulation,yao2022lassie}, or dense temporal correspondence in videos~\cite{yang2021viser,yang2021lasr,yang2021banmo}.
Without assuming any of these annotations/data, Hi-LASSIE introduces a novel and practical framework to estimate high-fidelity animal bodies by discovering 3D skeleton from a reference image.

%%%%%%%%%%%%%%%%%%%%%%%%%%%%%%%%%%%%%%%%%%%%%%%%%%%%%%%%%%%%
\vspace{1mm}
{\noindent  \textbf{Skeleton extraction and part discovery.}~}
2D skeleton/outline extraction has been widely studied and used as geometric context for shape/pattern recognition~\cite{belongie2002shape,xie2008shape,bai2008path,shen2016shape,he2022autolink}.
However, these methods often fail to identify separate skeleton bones/parts when they overlap or self-occlude each other in an image.
In this work, we propose to lift 2D skeletons to 3D for articulated shape learning by jointly considering the geometric and semantic cues in 2D images.
For part discovery, deep feature factorization (DFF)~\cite{collins2018deep} and follow-up works~\cite{hung2019scops,lathuiliere2020motion,choudhury2021unsupervised,amir2021deep} show that one could automatically obtain 2D corresponding part segments by clustering deep semantic features.
In the 3D domain, the object parts can be discovered by using volumetric cuboids~\cite{tulsiani2017learning}, clustering 3D point clouds~\cite{luo2020learning,paschalidou2020learning,paschalidou2021neural,mandikal20183d}, or learning part prior~\cite{yao2021discovering}.
%
%Tulsiani~\etal~\cite{tulsiani2017learning} use volumetric cuboids as part abstractions to learn 3D reconstruction.
%
% Mandikal~\etal~\cite{mandikal20183d} predict part-segmented 3D reconstructions from a single image.
% %
% In addition, Luo~\etal~\cite{luo2020learning} and  Paschalidou~\etal~\cite{paschalidou2020learning,paschalidou2021neural} learn to form object parts by clustering 3D points.
%
% Considering that most 3D approaches require ground-truth 3D shape of a whole object or its parts as supervision, Yao~\etal~\cite{yao2021discovering} propose to discover and reconstruct 3D parts automatically by learning a primitive shape prior.
%
These methods mainly assume some form of supervision like 3D shapes or camera viewpoints.
In this paper, we discover 3D parts of articulated animals from image ensembles without any 2D or 3D annotations/templates, which, to the best of our knowledge, is unexplored in the literature.

%%%%%%%%%%%%%%%%%%%%%%%%%%%%%%%%%%%%%%%%%%%%%%%%%%%%%%%%%%%%
\vspace{1mm}
{\noindent \textbf{3D reconstruction from sparse images.}~}
Optimizing a 3D scene or object from multi-view images is a fundamental problem in computer vision.
The majority of recent breakthroughs are based on the powerful Neural Radiance Field (NeRF)~\cite{mildenhall2020nerf} representation.
Given a set of multi-view images, NeRF learns a neural volume from which one can render high-quality novel views.
NeRS~\cite{zhang2021ners} introduces a neural surface representation to learn compact 3D shapes from a sparse image collection, which we find suitable to model individual animal body parts.
%
% Most NeRF-based frameworks assume multi-view images captured in the same illumination setting and that the object appearance (texture) remains the same across images.
%
Another line of work~\cite{boss2021nerd,zhang2021physg,boss2021neural} propose neural reflectance decomposition on image collections captured in varying illuminations, but they operate on rigid objects with ground-truth segmentation and known camera poses.
SAMURAI~\cite{boss2022samurai} jointly reasons about camera pose, shape, and materials from image collections of a single rigid object.
In contrast, our input consists of in-the-wild animal images with varying textures, viewpoints, and pose articulations captured in different environments.

\begin{figure*}[t!]
\centering
\includegraphics[width=.95\linewidth]{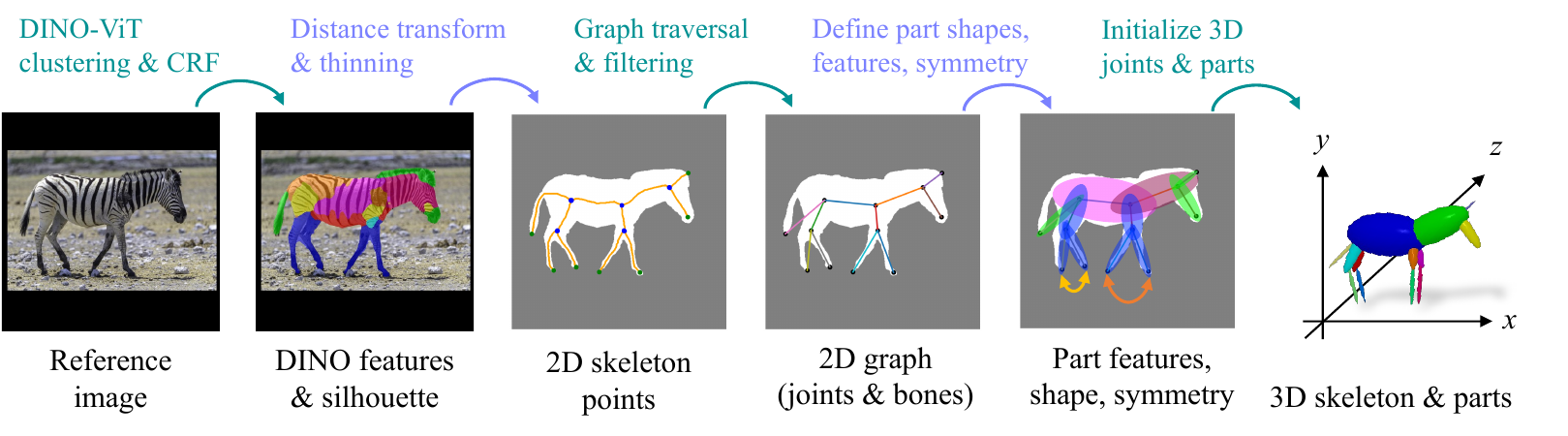}
\vspace{-3mm}
\caption{\textbf{3D Skeleton Discovery.} Given a reference image and its rough silhouette, we first extract, filter, and connect the 2D candidate points to form a 2D skeleton graph. Then, we find and split the symmetric parts by leveraging the 2D geometry and semantic cues. Finally, we design a simple heuristic to initialize the 3D joints, part shapes, and surface features.}
\vspace{-2mm}
\label{fig:skeleton}
\end{figure*}

\vspace{-2mm}
\section{Approach}
\vspace{-2mm}

Given a sparse image ensemble of an animal species, Hi-LASSIE first discovers a class-specific 3D skeleton that specifies the initial 3D joint coordinates and part connectivity.
Then, it jointly optimizes the camera viewpoint, pose articulation, and part shapes for each instance.
Before describing our approach, we briefly review the LASSIE~\cite{yao2022lassie} method, where several parts are adopted in this work.

% We briefly discuss a closely related work, LASSIE, in Section~\ref{sec:lassie}, then introduce our novel 3D skeleton discovery and optimization framework in Sections~\ref{sec:skeleton} and~\ref{sec:optimization}, respectively.

\vspace{-1mm}
\subsection{Preliminaries: 3D Skeleton \& Parts in LASSIE}
\vspace{-1mm}
\label{sec:lassie}
Given a user-provided 3D skeleton template that specifies the 3D joints and bones, LASSIE~\cite{yao2022lassie} represents an overall articulated shape by assembling several 3D parts.
In particular, each part is defined as a deformable neural surface~\cite{zhang2021ners} wrapped around a skeleton bone.
The part surfaces are parameterized by multi-layer perceptron networks (MLPs) which predict the deformation of any 3D point on a surface template (\eg unit sphere).
Formally, let $X \in \mathbb{R}^{3 \times m}$ denote the $m$ uniformly sampled 3D points on a unit sphere, one can deform the 3D surface of the $i$-th part in the canonical space using the part MLP as $X \mapsto \mathcal{F}_i(X)$.
The part surface is then rigidly transformed using the optimized 3D skeleton with scaling $s_i \in \mathbb{R}$, rotation $R_i \in \mathbb{R}^{3 \times 3}$, and translation $t_i \in \mathbb{R}^{3}$.
The final part surface points $V_i$ in the global coordinate frame can be expressed as:
\begin{equation}
    V_i = s_i R_i \mathcal{F}_i(X) + t_i . 
\label{eq:recon1}
\end{equation}
The rigid transformation of each part is defined by its corresponding bone length (scaling), orientation (rotation), and centroid (translation).
This skeleton-based representation ensures that the connectivity of 3D parts under arbitrary articulations and easy to repose.
Compared to explicit mesh representation, the neural surfaces also enable efficient mesh regularization while producing high-resolution surfaces since the MLPs take continuous surface coordinates as input.
A key innovation of LASSIE is the use of part prior within the part MLPs $\{\mathcal{F}_i\}$ that regularizes the part shapes to be close to convex primitive geometric shapes like spheres, cones, etc.
Please refer to~\cite{yao2022lassie} for further details.
To take advantage of these properties, we adopt a similar part-based representation in this work.
A key difference from LASSIE is that we alleviate the need for user-provided 3D skeleton template and instead discover the 3D skeleton automatically from a reference image chosen from the ensemble.
In addition, we propose several improvements to the neural surfaces and the optimization process to address the limitations of LASSIE resulting in higher-fidelity reconstructions.

%%%%%%%%%%%%%%%%%%%%%%%%%%%%%%%%%%%%%%%%%%%%%%%%%%%%%%%%%%%%
\vspace{-1mm}
\subsection{Discovering 3D Skeleton}
\vspace{-1mm}
\label{sec:skeleton}

% \iffalse
% \noindent {\bf Skeleton-based representation.}
% %
% Building an articulated shape from its 3D skeleton is an intuitive choice since all body parts are naturally grown upon and controlled by their bones and joints.
% %
% While an animal body can have a complicated shape, each part tends to have an uniform 3D shape and rigid motion. 
% %
% Moreover, part surfaces usually shared similar semantic features across instances.
% %
% These properties allow us to design effective pose and shape constraints to resolve certain 2D-3D ambiguities in images.
% %
% Similar to LASSIE, we adopt a part-based representation built upon a 3D skeleton that defines the joint coordinates at resting pose and their connectivity (bones).
% %
% As such, an animal body can be formed by assembling the 3D part surfaces around each bone after instance-specific bone transformations like scaling and rotations.
% %
% Nonetheless, instead of relying on a human-annotated 3D skeleton, we further reduce the need of user intervention by discovering 3D skeleton automatically.
% %
% The only manual input is to select a reference image from the ensemble, where most body parts are visible.
% \fi

% \vspace{2mm}
\noindent {\bf 2D skeleton extraction.}
Fig.~\ref{fig:skeleton} illustrates our 3D skeleton discovery process from a reference image.
We first extract a set of 2D skeleton points from a reference silhouette using a skeletonize/morphological thinning algorithm~\cite{zhang1984fast}.
Specifically, we iteratively remove pixels from the borders until none can be removed without altering the connectivity.
These 2D points can be seen as candidates of underlying skeleton joints and bones through camera projection.
Similar to prior 2D shape matching methods~\cite{xie2008shape,bai2008path}, we categorize the skeleton points into junctions, endpoints, and connection points.
Assuming that each skeleton curve is one pixel wide, a skeleton point is called an \textit{endpoint} if it has only one adjacent point; a \textit{junction} if it has three or more adjacent points; and a \textit{connection point} if it is neither an endpoint nor a junction.
We sort the junctions and endpoints by distance transform (2D distance to closest border) and identify the `root' junction with the largest distance to the border.
Next, we find the shortest path (sequence of skeleton points) between root and each endpoint by traversing though the graph.
From these paths, we can build a skeleton tree with 2D joints (junctions or endpoints) and bones (junction-junction connection or junction-endpoint connection).
Since the silhouettes and skeleton points are sometimes noisy, we filter those noisy 2D joints using non-maximal suppression.
That is, we remove a point is if it lies within the coverage (radius calculated by distance transform) of its parent joint.

\begin{figure*}[t!]
\centering
\includegraphics[width=.95\linewidth]{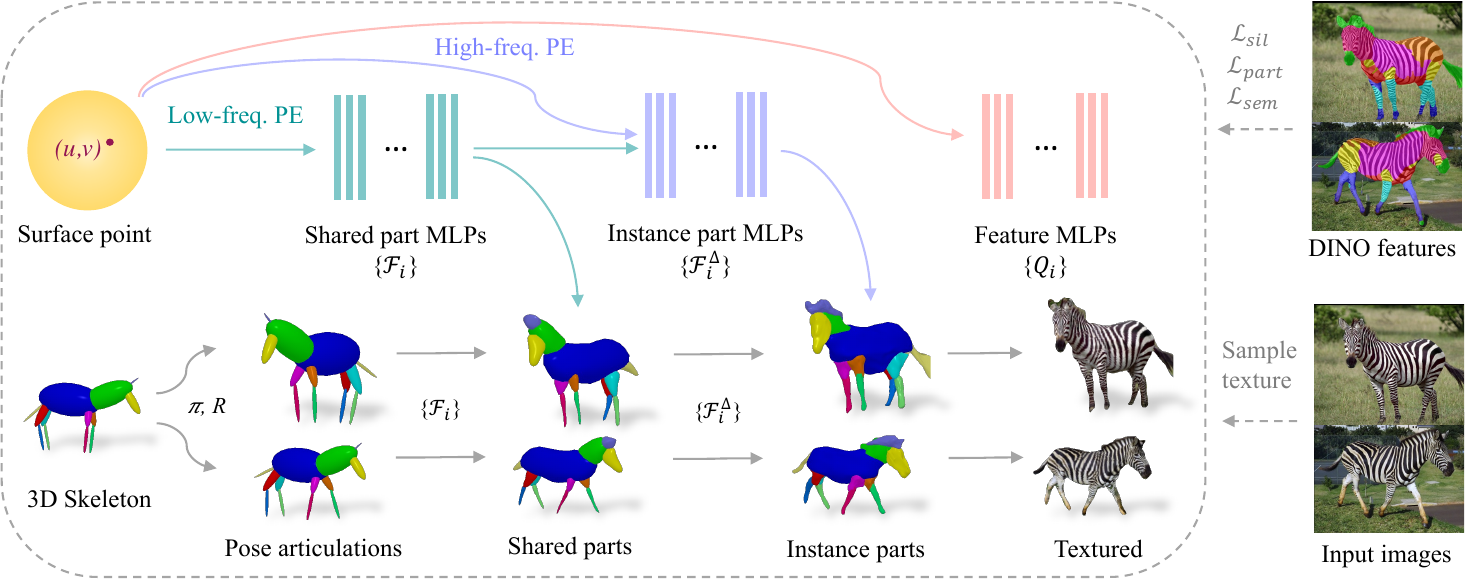}
\vspace{-3mm}
\caption{\textbf{Hi-LASSIE optimization framework.} Based on the discovered 3D skeleton, we reconstruct an articulated shape by optimizing the camera viewpoints, pose articulations, and part shapes. We represent the 3D part shapes as neural surfaces, which are decomposed into shared (low-frequency) and instance-specific (high-frequency) components via positional encoding (PE) of input surface coordinates. The only image-level annotation is from the self-supervisory DINO features, and the surface texture is sampled from the input images.}
\vspace{-2mm}
\label{fig:framework}
\end{figure*}

\vspace{1mm}
\noindent {\bf 3D joint and part initialization.}
From the extracted 2D skeleton, one can roughly initialize the simple part shapes and surface features using the reference image.
However, all the joints and bones would be initialized on a 2D plane (image plane) which is not sufficient for later 3D shape and camera pose optimization across instances.
% without knowing the depth value of individual parts, which often results in sub-optimal output shapes.
%
To obtain a better 3D skeleton initialization, we propose to find symmetric parts and separate them in the 3D space w.r.t. the symmetry plane.
Our key insight is that symmetric parts (\eg left and right legs/ears) share similar geometric and semantic features.
Therefore, we design a heuristic to calculate the symmetry score of each pair of joints/bones based on the their geometric and semantic feature distance.
Specifically, we compute the following features for each joint/bone: length, average radius, and the average DINO features~\cite{caron2021emerging} along the paths to their common ancestor in the skeleton tree.
Then, we identify pairs of joints that share similar features, which usually correspond to the symmetric animal parts (e.g., left and right leg).
% If two joints are similar both geometrically and semantically (in terms of feature distance), 
To uplift our 2D skeleton onto 3D, we set the z-coordinates of the each joint pairs to be on opposite sides of the symmetry plane ($z=0$) and offset by their average radius (so that the initial 3D parts do not overlap).
To better deal with overlapping parts in 2D silhouettes, we also split the parent of two symmetric parts if it has only two children.
See Fig.~\ref{fig:skeleton} (right) for an example of estimated 3D skeleton.

%%%%%%%%%%%%%%%%%%%%%%%%%%%%%%%%%%%%%%%%%%%%%%%%%%%%%%%%%%%%
\vspace{-1mm}
\subsection{Learning High-fidelity Articulated Shapes}
\vspace{-1mm}
\label{sec:optimization}

\noindent {\bf Optimization setting.}
Our optimization framework takes a sparse set of $n$ (typically 20-30) in-the-wild images $\{I_j\}_{j=1}^n$ as well as the discovered 3D skeleton $P\in\mathbb{R}^{p\times 3}$ with $p$ joints and $b$ bones/parts as input.
The only image-level annotations like silhouettes and semantic clusters are obtained from the clustering results of self-supervisory DINO~\cite{caron2021emerging} features.
All the images in an ensemble contain instances of the same species, but each instance could vary in pose, shape, texture, camera viewpoint, etc.
We assume that each animal body is mostly visible in an image without severe truncation or occlusions by other objects although self-occlusion is allowed.
We use $j$ to denote the index over images $j \in \{1,.,n\}$ and $i$ the index over parts $i \in \{1,.,b\}$.
For each instance, Hi-LASSIE optimizes the camera viewpoint $\pi^j=(R_0,t_0)$, part rotations $R^j\in\mathbb{R}^{b \times 3 \times 3}$, and neural part deformation MLPs.
The overall framework is shown in Fig.~\ref{fig:framework}, where we first perform pose articulation then progressively add more geometric details (from the shared part shapes to instance-specific deformations).

\begin{figure}[t!]
\centering
\includegraphics[width=.95\linewidth]{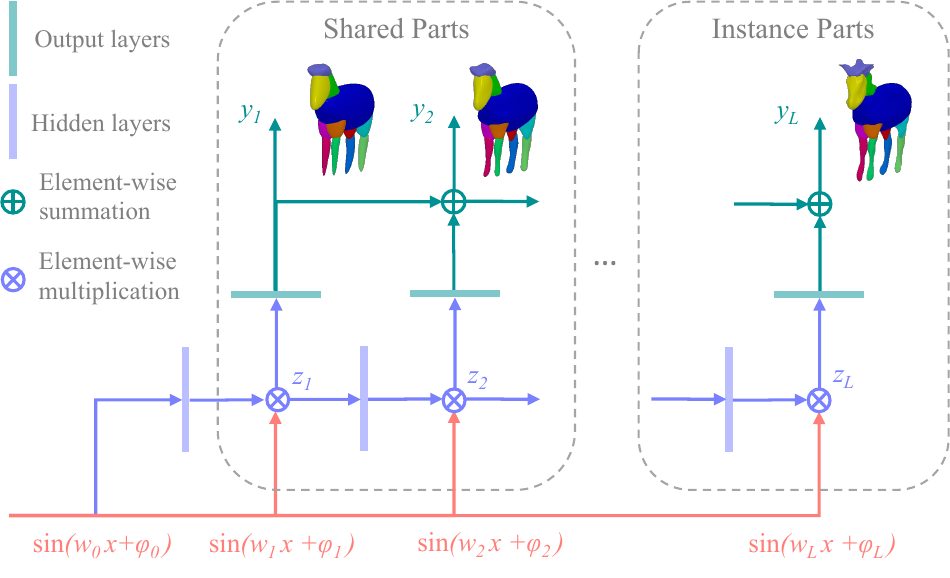}
\vspace{-2mm}
\caption{\textbf{Frequency-decomposed part MLP for per-instance shape deformation.} We design a MLP network to represent the neural surfaces composed of varying amount of details. By increasing the frequencies of input positional encoding, the outputs of deeper layers can express more detailed deformation. The output and hidden layers are linear layers.}
\vspace{-2mm}
\label{fig:frequency}
\end{figure}

\vspace{1mm}
\noindent {\bf Per-instance deformation by frequency decomposition.}
Unlike LASSIE where the part shapes are shared across all instances, which limits the shape fidelity to input images, we propose to learn instance-specific part shapes that can better account for instance-varying and articulation-dependent deformations.
Intuitively, instances of the same species should share similar base part shapes and only deviate in high-frequency details (\eg ears and tail).
Such detailed difference is usually caused by part articulations or instance variance.
We observe that naively fine-tuning the part surfaces on one instance tends to overfit to that instance and nullify the 3D shape prior learned across all instances.
For instance, the part shapes can overfit to the target silhouette (input view) but look unrealistic in novel views.
To address this, we propose to decompose the surface deformations in the spatial frequency space so that each instance can have high-frequency detail variations while sharing the same low-frequency base part shapes.
As shown in Fig.~\ref{fig:frequency}, we achieve this by designing a part deformation MLP with frequency-decomposed input and output components.
In a forward pass through the MLP, we first apply positional encoding (PE) to an input coordinate $x\in\mathbb{R}^3$ using sines with varying frequencies $\omega$ and phases $\phi$: $\text{PE}_i$($x$) $= \text{sin}(\omega_i x + \phi_i)$, where $i=0,...,L$, and $L$ is the number of layers in the network.
The hidden activations $z_i$ and final outputs $y_i$ of layer $i$ are computed as:
\begin{align}
    z_i &= \text{PE}_i(x) \otimes (W_i^\text{h} z_{i-1} + b_i^\text{h}) \mathrm{~with~} z_0 = \text{PE}_0(x);  \\
    y_i &= y_{i-1} + (W_i^\text{o} z_i + b_i^\text{o}) \mathrm{~with~} y_0 = \mathbf{0}; 
\label{eq:bacon}
\end{align}
where $\otimes$ denotes the Hadamard (element-wise) product, ($W^\text{h},b^\text{h}$) parametrize the hidden linear layers, and ($W^\text{o},b^\text{o}$) represent the output linear layers.
This formulation leverages a useful property that 
repeated Hadamard product of sines are equivalent to a sum of sines with varying amplitude, frequency, and phase~\cite{fathony2020multiplicative}.
Therefore, with increasing PE frequencies from shallow to deep layers, we can produce output shapes with increasing amount of high-frequency details.
This allows us to perform instance-specific optimization while preserving the common base shapes by sharing the shallow layers of part MLPs across instances and optimizing the deep layers to be instance-specific as shown in Fig.~\ref{fig:frequency}.
Note that our MLP network is inspired from band-limited coordinate networks (BACON)~\cite{lindell2022bacon}.
However, we add the cumulative sum of all previous outputs to the current output of each layer since the high-frequencies details should be deformations from low-frequency based shapes.
Moreover, unlike BACON, that learns the frequencies of sine functions in PE together with other parameters, we pre-define and fix the input frequencies to improve the optimization stability in our ill-posed problem and better control the separation of shared and instance components.

\begin{figure}[t!]
\centering
\includegraphics[width=.95\linewidth]{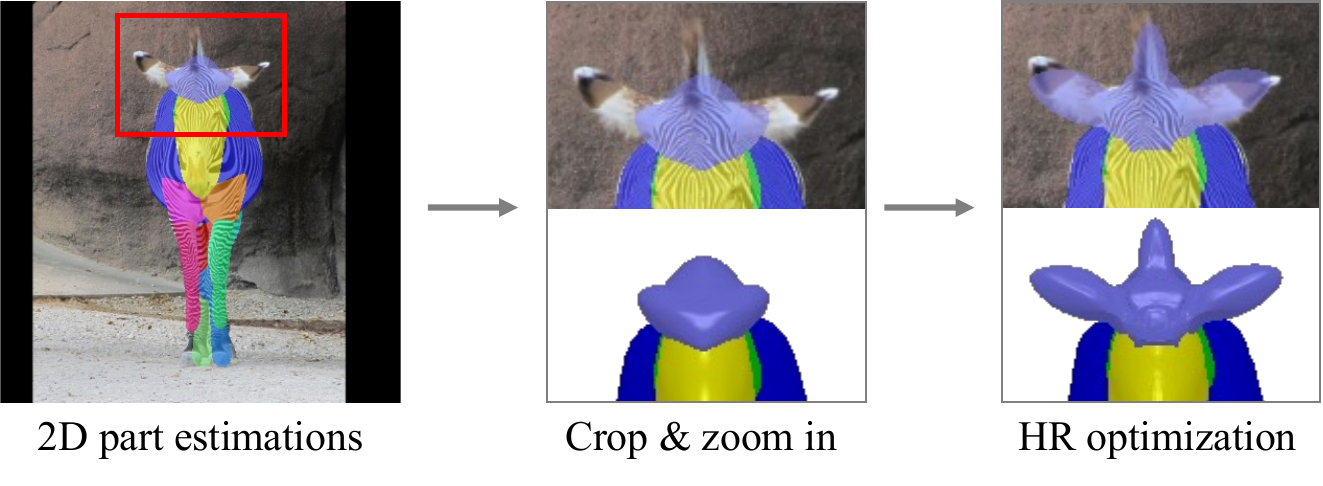}
\vspace{-2mm}
\caption{\textbf{High-resolution rendering and optimization by zooming in on parts.} Based on the initial estimates of 2D part localization, we crop and upsample each part region to perform shape optimization at higher resolution.}
\vspace{-2mm}
\label{fig:zoomin}
\end{figure}

\vspace{1mm}
\noindent {\bf Zoom in to parts for higher details}. % \& silhouette losses.}
Similar to LASSIE, we perform analysis-by-synthesis to supervise the overall shape reconstruction since we do not have access to any form of 3D supervision.
We render the part surfaces via a differentiable renderer~\cite{liu2019soft} and compare them with the pseudo ground-truth masks obtained from DINO feature clustering.
The silhouette loss $\mathcal{L}_{sil}$ is written as:
$\mathcal{L}_{sil} = \sum_j \lVert  M^j - \hat{M}^j \rVert^2$ ,
where $M^j$ and $\hat{M}^j$ are the rendered silhouette and pseudo ground-truth of instance $j$, respectively.
To capture more shape details in the images, we propose to render and compare the high-resolution silhouettes by zooming in on individual parts during optimization.
Concretely, we crop the 2D part masks estimated by Hi-LASSIE, upsample the cropped regions to higher resolution, and calculate the part silhouette loss $\mathcal{L}_{part}$ on the zoomed-in part masks as:
\begin{align}
    \mathcal{L}_{part} = \sum_i \sum_j \Big\lVert  {\Gamma_i}^j(M^j) - {\Gamma_i}^j(\hat{M}^j) \Big\rVert^2,
\label{eq:part}
\end{align}
where ${\Gamma_i}^j$ denotes the crop-and-upsample operation for part $i$ on instance $j$.
We show some example outputs before and after zoomed-in optimization in Fig.~\ref{fig:zoomin}.

\vspace{2mm}
\noindent {\bf 2D-3D semantic consistency via feature MLPs.}
To densely enforce the 2D-3D semantic consistency of part surfaces, we further introduce feature MLPs to improve the semantic loss proposed in LASSIE.
For each instance $j$, the semantic consistency loss is defined as the Chamfer distance between the foreground pixels $\{p|\hat{M}^j(p)=1\}$ and 3D surface points $\{x\in X\}$ in a high-dimensional space:
\begin{equation}
    \mathcal{L}_{sem} = \sum\limits_{j} \Big( \sum\limits_{p} \hspace{1mm} \min\limits_{x} \hspace{1mm} \mathcal{D}(p,x) + \sum\limits_{x} \hspace{1mm} \min\limits_{p} \hspace{1mm} \mathcal{D}(p,x) \Big) .
\label{eq:chamfer}
\end{equation}
The high-dimensional distance $\mathcal{D}$ is defined as:
\begin{equation}
    \mathcal{D}(p,x) = \underbrace{\lVert \pi^j(x) - p \rVert^2}_\text{Geometric distance} \hspace{1mm} + \hspace{1mm} \alpha \underbrace{\lVert Q(x) - K^j(p) \rVert^2}_\text{Semantic distance} ,
\label{eq:dist}
\end{equation}
where $\alpha$ is a scalar weighting for semantic distance, $Q(x)$ denotes the 3D surface features of point $x$, and $K(p)$ is the 2D image features at pixel $p$. 
In effect, $\mathcal{L}_{sem}$ optimizes the 3D surface coordinates such that the aggregated 3D point features would project closer to the similar pixel features in the image ensemble.
In LASSIE, the 3D surface features are maintained and updated for only a sparse set of surface points, which limits the capabilities of $\mathcal{L}_{sem}$ to supervise part localization.
Instead, we represent the 3D surface feature of each part with an MLP (Fig~\ref{fig:framework}), which is similar to the part shape MLPs shared by all instances.
Consequently, we can obtain the semantic features $Q(x)$ given an arbitrary surface coordinate $x$.
We update the part surface and feature MLPs alternatively in an EM-style optimization.
That is, we update the feature MLPs by sampling 3D surface points and projecting them onto 2D images in the E-step.
In the M-step, we use the updated features to optimize 3D surface MLPs via minimizing the semantic consistency loss.

\vspace{1mm}
\noindent {\bf Pose and shape regularizations.}
To constrain the output articulations and part shapes, we apply the following pose and shape regularizations.
First, we impose a part rotation loss $\mathcal{L}_{rot}$ to limit the angle offsets from resting pose as:
$\mathcal{L}_{rot} = \sum_j \lVert R^j - \bar{R} \rVert^2$,
where $R^j$ is the part rotations of instance $j$ and $\bar{R}$ denotes the part rotations of shared resting pose.
Moreover, we remove the side-way rotation constraint on animal legs in LASSIE since the leg parts are not specified in our self-discovered skeletons.
Instead, we propose a more general regularization based on 3D symmetry prior.
We define the symmetry loss $\mathcal{L}_{sym}$ on the 3D joints to prevent overlapping parts or irregular poses as:
$\mathcal{L}_{sym} = \sum_i \lVert J_i - \Psi({J_i}^\star) \rVert^2$
where ${J_i}^\star$ is the symmetric joint of $J_i$ and $\Psi$ is the reflection operation w.r.t. the symmetry plane.
Finally, to encourage smooth part surfaces, we apply common mesh regularizations like Laplacian loss $\mathcal{L}_{lap}$ and surface normal loss $\mathcal{L}_{norm}$.
$\mathcal{L}_{lap}$ encourages smooth 3D surfaces by pulling each vertex towards the center of its neighbors, and $\mathcal{L}_{norm}$ enforces neighboring mesh faces to have similar normal vectors.
Note that all the pose and shape regularizations are generic and applicable to a wide range of articulated shapes.

\vspace{1mm}
\noindent {\bf Optimization and texture sampling.}
The overall optimization objective is given by the weighted sum of aforementioned losses ($\mathcal{L}_{sil}$, $\mathcal{L}_{part}$, $\mathcal{L}_{sem}$, $\mathcal{L}_{rot}$, $\mathcal{L}_{sym}$, $\mathcal{L}_{lap}$, $\mathcal{L}_{norm}$).
We optimize the camera, pose, and shape parameters in a multi-stage manner.
Superficially, we first estimate the camera viewpoints and fix the rest.
Then, we optimize the part transformations and shared part MLPs along with cameras until convergence.
Finally, we freeze shared part MLPs (shallow layers) and fine-tune the instance-specific part deformations (deep layers) on each instance individually.
Note that the surface feature MLPs are also updated during all optimization stages in an EM-style.
The final textured outputs are generated by densely sampling the colors of visible surface points from individual images.
The invisible (self-occluded) surfaces, on the other hand, are textured by their symmetric surfaces or nearest visible neighbors.
More details can be found in the supplemental material.

\begin{figure*}[t]
%
% \vspace{-3mm}
\centering
\small
\setlength\tabcolsep{0pt}
\renewcommand{\arraystretch}{0.0}
\begin{tabular}[t!]{cc ccccc cc}
%\includegraphics[width=.11\linewidth]{figures/results/zebra/proc_3.png} &
%\includegraphics[width=.11\linewidth]{figures/results/zebra/acsm_view0_1.png} &
%\includegraphics[width=.11\linewidth]{figures/results/zebra/part_1.png} &
%\includegraphics[width=.11\linewidth]{figures/results/zebra/opt_inst_part_3.png} &
%\includegraphics[width=.11\linewidth]{figures/results/zebra/opt_inst_view_part_3_30.png} &
%\includegraphics[width=.11\linewidth]{figures/results/zebra/opt_inst_text_3.png} &
%\includegraphics[width=.11\linewidth]{figures/results/zebra/opt_inst_view_text_3_30.png} &
%\includegraphics[width=.11\linewidth]{figures/results/zebra/opt_global_part_28.png}
% \\
\includegraphics[width=.11\linewidth]{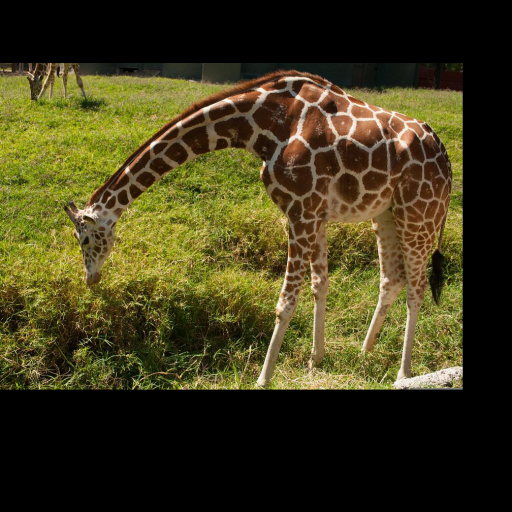} &
\includegraphics[width=.11\linewidth]{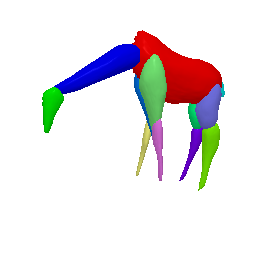} &
\includegraphics[width=.11\linewidth]{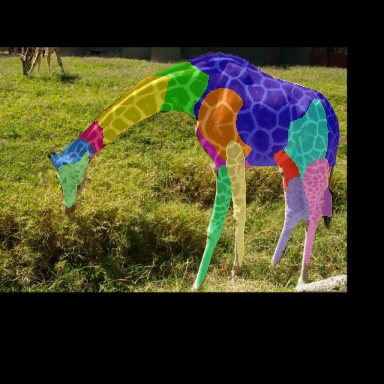} &
\includegraphics[width=.11\linewidth]{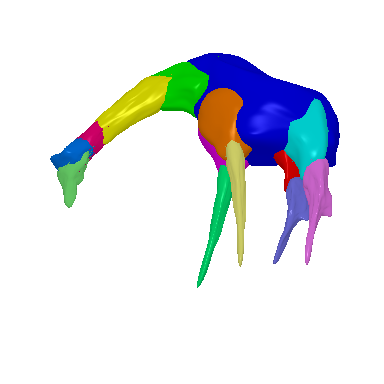} &
\includegraphics[width=.11\linewidth]{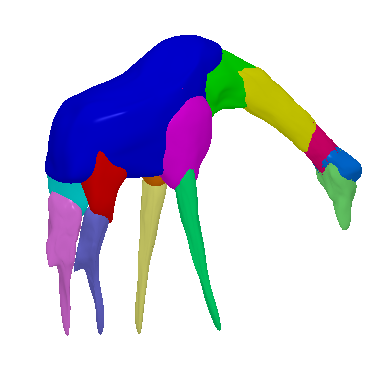} &
\includegraphics[width=.11\linewidth]{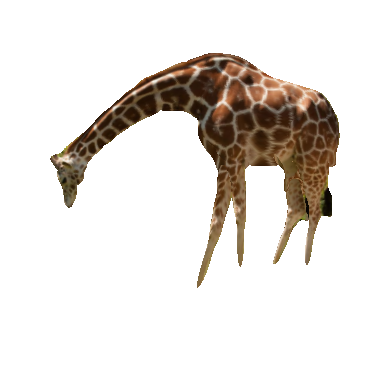} &
\includegraphics[width=.11\linewidth]{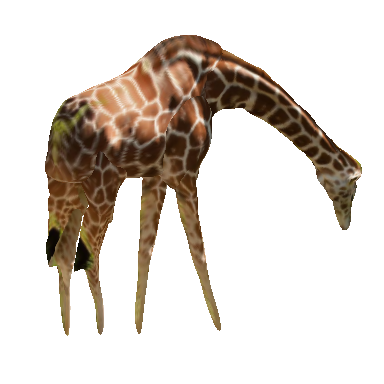} &
\includegraphics[width=.11\linewidth]{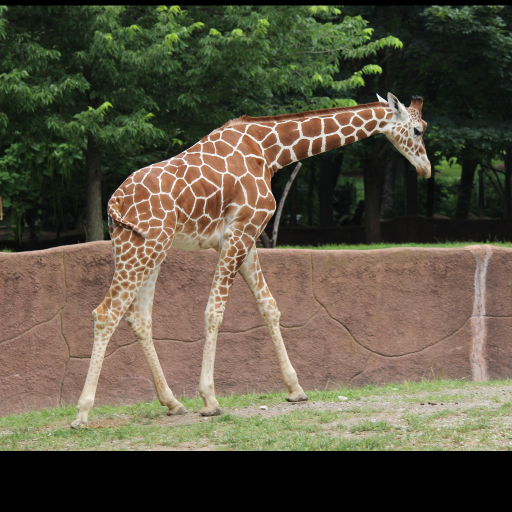} &
\includegraphics[width=.11\linewidth]{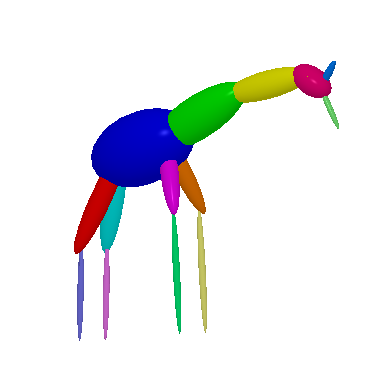}
\vspace{-2mm}
\\
\includegraphics[width=.11\linewidth]{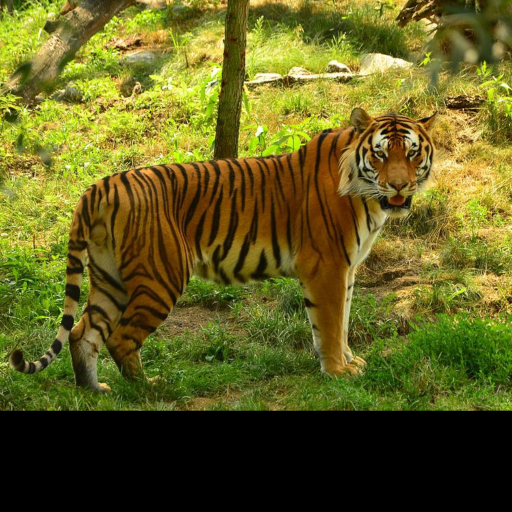} &
\includegraphics[width=.11\linewidth]{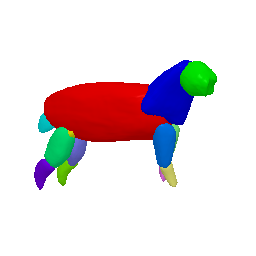} &
\includegraphics[width=.11\linewidth]{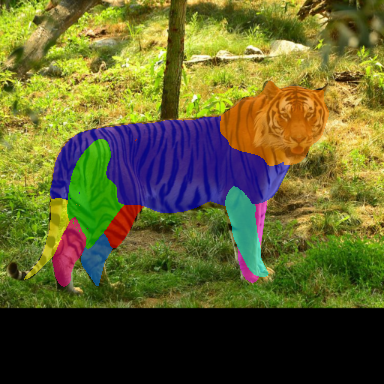}  &
\includegraphics[width=.11\linewidth]{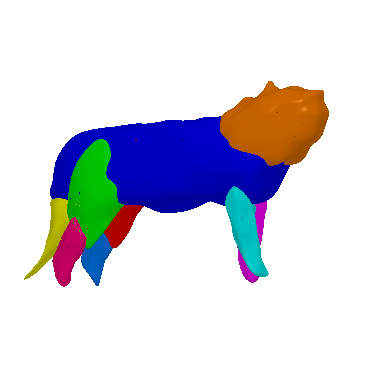} &
\includegraphics[width=.11\linewidth]{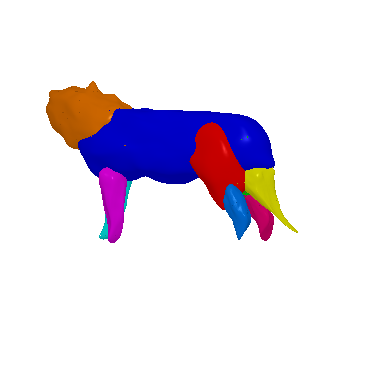} &
\includegraphics[width=.11\linewidth]{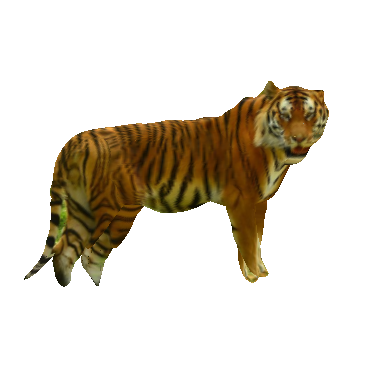} &
\includegraphics[width=.11\linewidth]{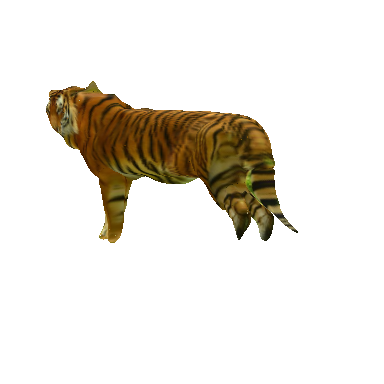} &
\includegraphics[width=.11\linewidth]{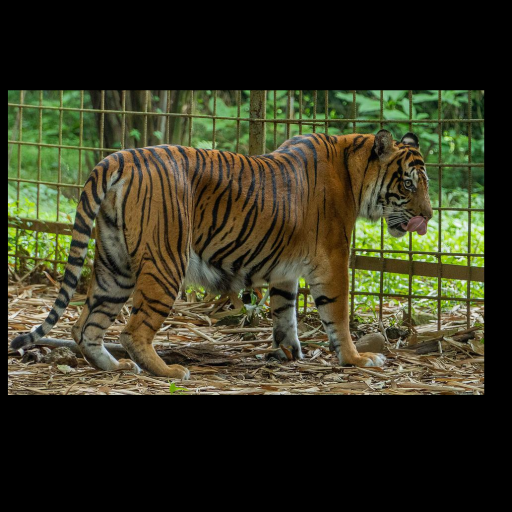} &
\includegraphics[width=.11\linewidth]{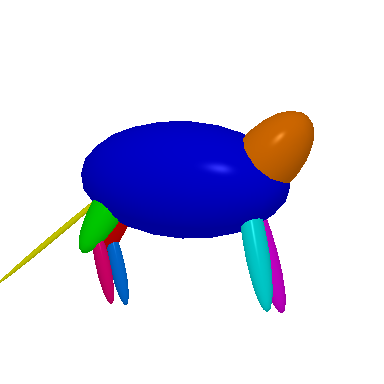}
\vspace{-2mm}
\\
\includegraphics[width=.11\linewidth]{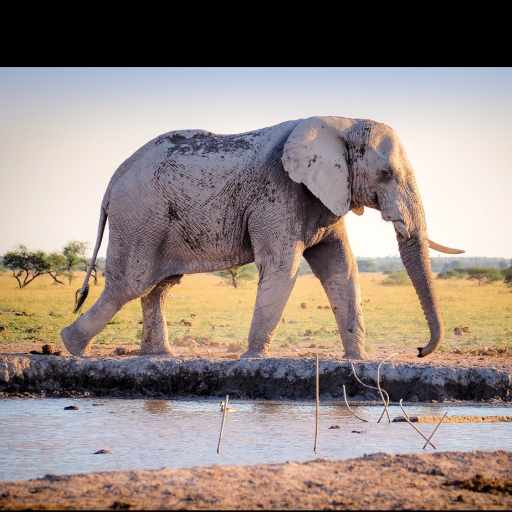} &
\includegraphics[width=.11\linewidth]{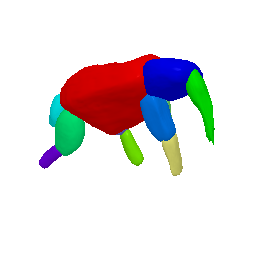} &
\includegraphics[width=.11\linewidth]{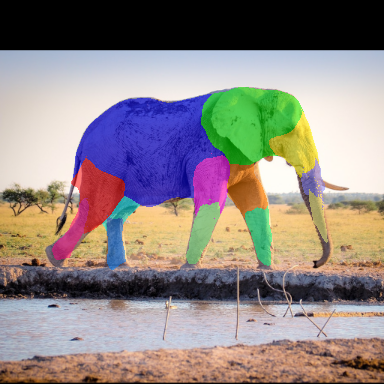} &
\includegraphics[width=.11\linewidth]{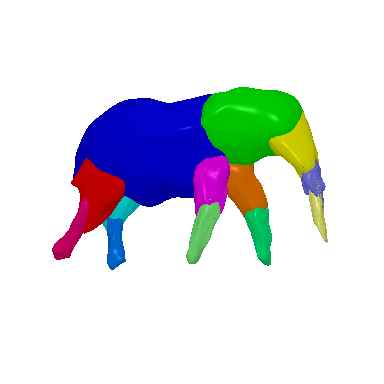} &
\includegraphics[width=.11\linewidth]{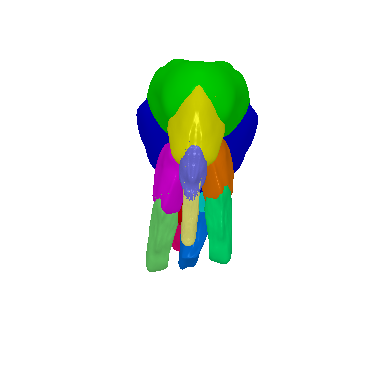} &
\includegraphics[width=.11\linewidth]{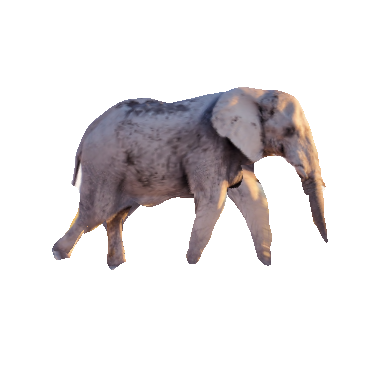} &
\includegraphics[width=.11\linewidth]{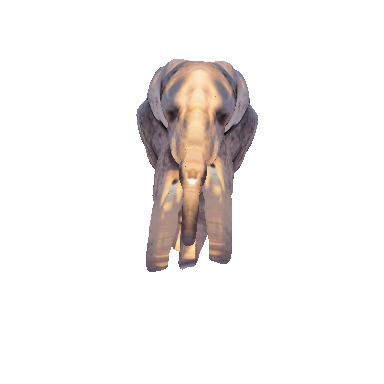} &
\includegraphics[width=.11\linewidth]{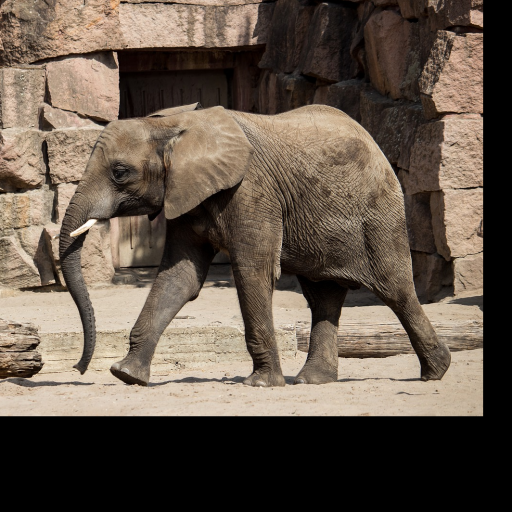} &
\includegraphics[width=.11\linewidth]{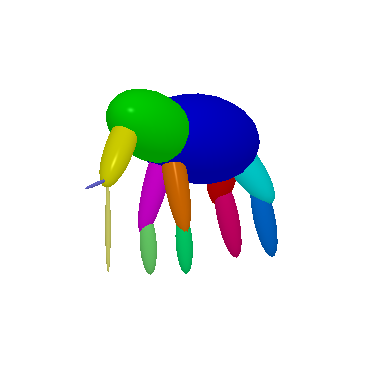}
\vspace{-2mm}
\\
\includegraphics[width=.11\linewidth]{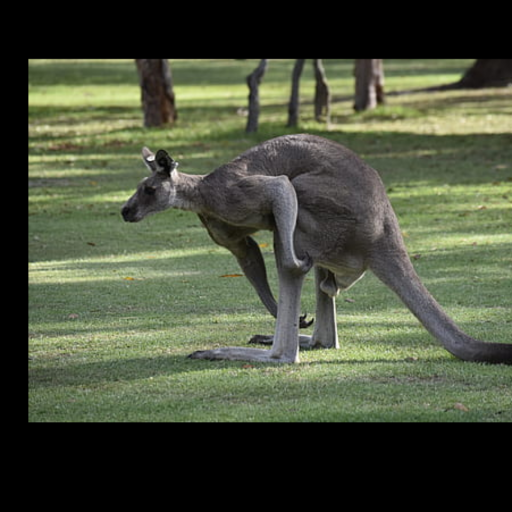} &
\includegraphics[width=.11\linewidth]{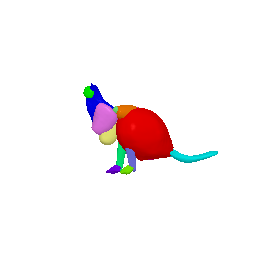} &
\includegraphics[width=.11\linewidth]{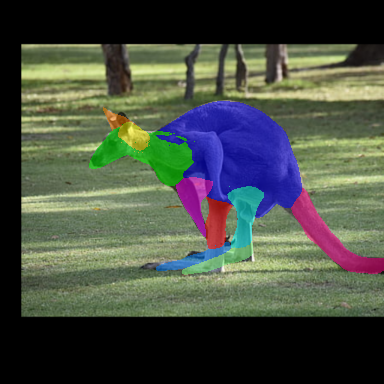} &
\includegraphics[width=.11\linewidth]{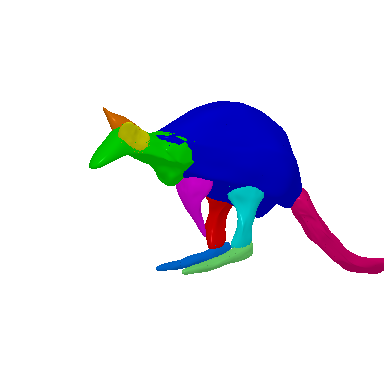} &
\includegraphics[width=.11\linewidth]{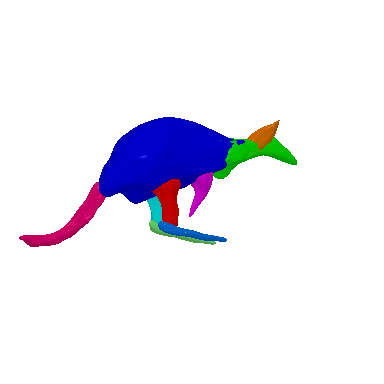} &
\includegraphics[width=.11\linewidth]{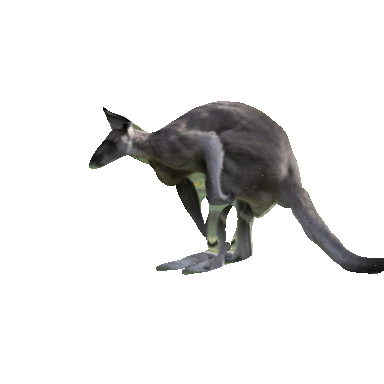} &
\includegraphics[width=.11\linewidth]{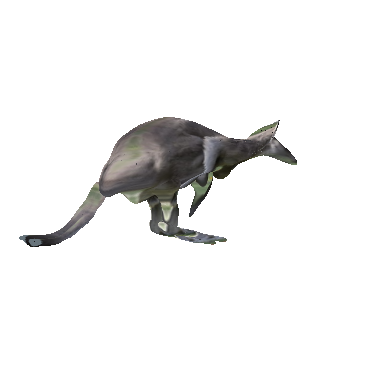} &
\includegraphics[width=.11\linewidth]{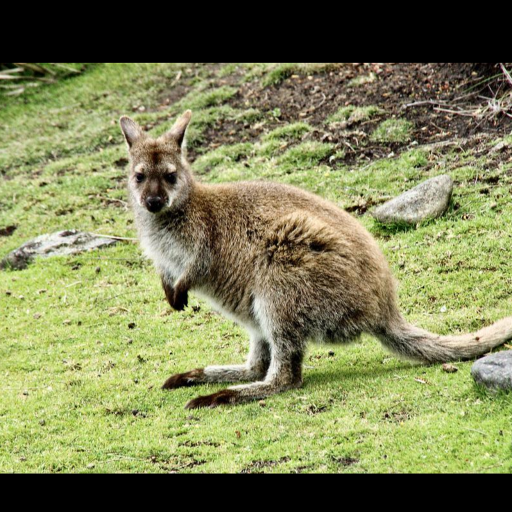} &
\includegraphics[width=.11\linewidth]{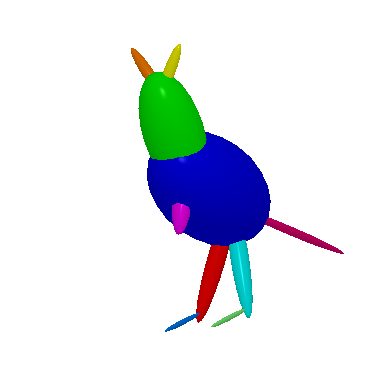}
\vspace{-2mm}
\\
\includegraphics[width=.11\linewidth]{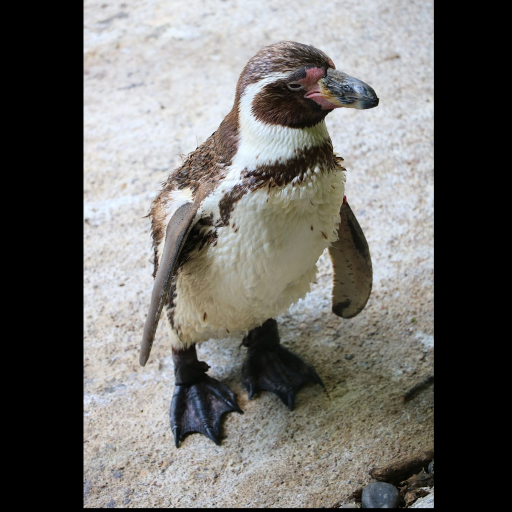} &
\includegraphics[width=.11\linewidth]{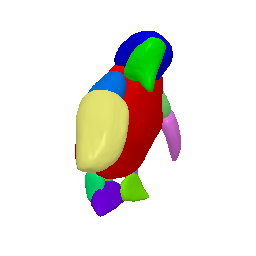} &
\includegraphics[width=.11\linewidth]{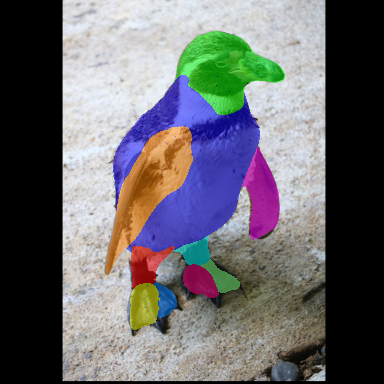} &
\includegraphics[width=.11\linewidth]{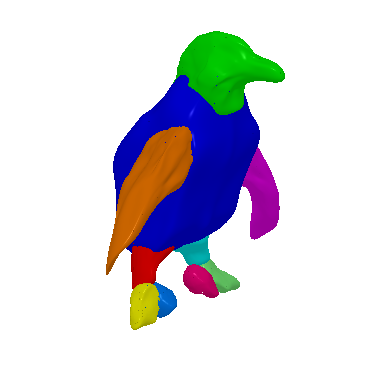} &
\includegraphics[width=.11\linewidth]{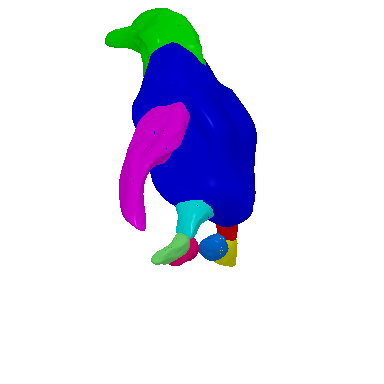} &
\includegraphics[width=.11\linewidth]{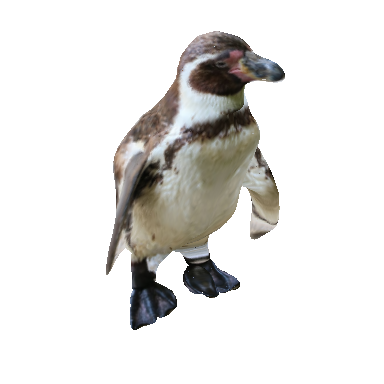} &
\includegraphics[width=.11\linewidth]{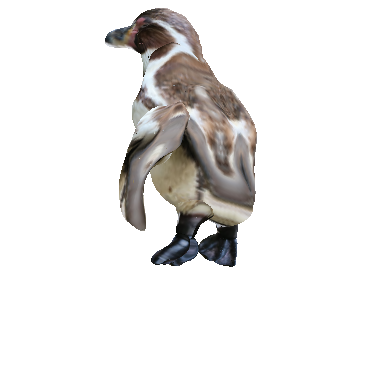} &
\includegraphics[width=.11\linewidth]{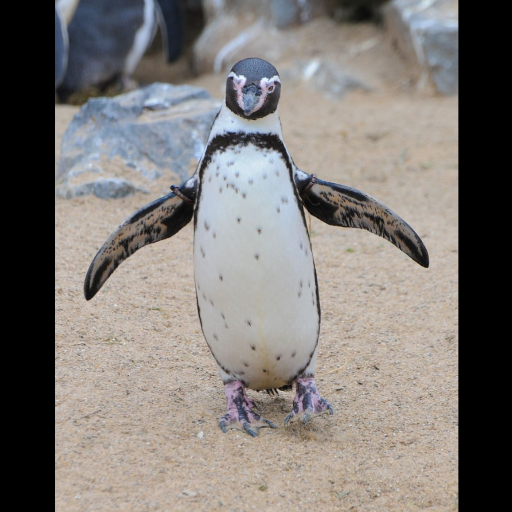} &
\includegraphics[width=.11\linewidth]{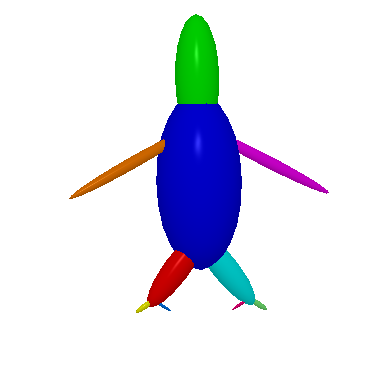}
\vspace{2mm}
\\
Input & % A-CSM~\cite{kulkarni2020articulation} & 
LASSIE~\cite{yao2022lassie} & \multicolumn{3}{c}{Hi-LASSIE (2D/3D parts)} & \multicolumn{2}{c}{Hi-LASSIE (texture)} & Reference & Skeleton
\end{tabular}
\vspace{-2mm}
\caption{\textbf{Qualitative results on in-the-wild images.} We show example results of prior arts and Hi-LASSIE on the LASSIE~\cite{yao2022lassie} animal image ensembles as well as the reference image and 3D skeleton discovered by Hi-LASSIE.  The results demonstrate the effectiveness of our 3D skeleton discovery and the high-fidelity shape/texture reconstruction across diverse animal classes.}
\label{fig:all}
\vspace{-2mm}
\end{figure*}

\vspace{2mm}
\section{Experiments}
\vspace{-2mm}

%%%%%%%%%%%%%%%%%%%%%%%%%%%%%%%%%%%%%%%%%%%%%%%%%%%%%%%%%%%%
{\noindent \textbf{Datasets and baselines.}~}
We follow the same evaluation protocols as in LASSIE~\cite{yao2022lassie}.
That is, we optimize and evaluate Hi-LASSIE on individual image ensembles in the Pascal-Part~\cite{chen2014detect} and LASSIE~\cite{yao2022lassie} datasets.
From the Pascal-Part dataset, we select the same set of images of horse, cow, and sheep as in LASSIE.
These image are annotated with 2D part segmentation masks, which we use to automatically find 2D keypoints for evaluation. 
The LASSIE dataset contains sparse image ensembles (with CC-licensed web images) of some diverse animals (zebra, tiger, giraffe, elephant, kangaroo, and penguin) with 2D keypoints annotations for evaluation.
Considering the novelty of this problem setting (sparse image optimization for articulated animal shapes), we mainly compare Hi-LASSIE with LASSIE as well as some learning-based mesh reconstruction methods.
Most recent mesh reconstruction methods either cannot handle articulations~\cite{kulkarni2019canonical,li2020self,goel2020shape,tulsiani2020implicit,ye2021shelf} or leverage different inputs~\cite{li2020online,yang2021lasr,yang2021banmo}.
3D Safari~\cite{zuffi2019three} and A-CSM~\cite{kulkarni2020articulation}, on the other hand, are more comparable to Hi-LASSIE since they explicitly model articulations for animal classes of our interest.
Since both 3D Safari and A-CSM require large-scale image sets for training (not available in our setting), we use their released models of the closest animal classes to evaluate on our datasets. 
For instance, we use the zebra model of 3D Safari and the horse model of A-CSM to evaluate on similar quadrupeds.

%%%%%%%%%%%%%%%%%%%%%%%%%%%%%%%%%%%%%%%%%%%%%%%%%%%%%%%%%%%%
\vspace{1mm}
{\noindent \textbf{Visual comparisons.}~}
Fig.~\ref{fig:all} shows the qualitative results of Hi-LASSIE and prior methods on LASSIE dataset images.
%
% A-CSM~\cite{kulkarni2020articulation} assumes a 3D template mesh and class-specific surface maps, which enable detailed shapes of animal bodies.
%
% However, its outputs do not align well with the given 2D images.
%
LASSIE results can fit the object silhouettes quite well but lack shape details.
Our results demonstrate that Hi-LASSIE can effectively discover a good 3D skeleton that well explains other instance in the image ensemble.
Moreover, the output articulated parts are detailed, high-fidelity, and faithful to input images.
More qualitative results of skeleton and shape discovery are shown in the supplemental material.
%
% Note that our 3D reconstructions also improve the 2D part segmentation from DINO-ViT feature clustering~\cite{amir2021deep}.

\begin{table*}[t!]
\centering
\caption{\textbf{Keypoint transfer evaluations on the Pascal-Part~\cite{chen2014detect} and LASSIE~\cite{yao2022lassie} image ensembles.} For all pairs of images in each animal class, we report the average percentage of correct keypoints (PCK@0.05).}
\vspace{-2mm}
\small
\begin{tabular}{l ccccccccc}
\toprule
    % & \multicolumn{3}{c}{Pascal-Part dataset} & \multicolumn{6}{c}{LASSIE dataset} \\
    % \cmidrule(lr){2-4}\cmidrule(lr){5-10}
    Method & Horse & Cow & Sheep & Zebra & Tiger & Giraffe & Elephant & Kangaroo & Penguin
    \\
\midrule
    3D Safari~\cite{zuffi2019three} & 57.1 & 50.3 & 50.5 & 62.1 & 50.3 & 32.5 & 29.9 & 20.7 & 28.9
    \\
    A-CSM~\cite{kulkarni2020articulation} & 55.3 & 60.5 & 54.7 & 60.3 & 55.7 & 52.2 & 39.5 & 26.9 & 33.0
    \\
    LASSIE~\cite{yao2022lassie} & 58.0 & 62.4 & 55.5 & 63.3 & 62.4 & 60.5 & 40.3 & 31.5 & 40.6
    \\
\midrule
    Hi-LASSIE w/o inst. MLPs & 56.8 & 57.7 & 53.6 & 59.7 & 63.0 & 59.9 & 40.8 & 31.5 & 41.2
    \\
    Hi-LASSIE w/o feat. MLPs & 57.5 & 62.2 & \bf56.3 & 64.1 & 62.0 & 60.8 & \bf42.7 & 30.3 & 41.5
    \\
    Hi-LASSIE w/o $\mathcal{L}_{part}$ & 58.8 & 62.8 & 55.9 & 63.8 & 62.8 & 61.1 & 41.5 & 33.3 & 42.5
    \\
    Hi-LASSIE & \bf59.6 & \bf63.1 & 56.2 & \bf64.2 & \bf63.1 & \bf61.6 & \bf42.7 & \bf35.0 & \bf44.4
    \\
\bottomrule
\end{tabular}
\label{tab:keypoint}
% \vspace{-4mm}
\end{table*}

\begin{table*}[tp]
\centering
\caption{\textbf{Quantitative evaluations on the Pascal-part images.} We report the overall 2D IOU, part mask IOU, and percentage of correct pixels (PCP) under dense part mask transfer between all source-target image pairs.}
\vspace{-2mm}
\small
\setlength\tabcolsep{8pt}
\begin{tabular}{l ccc ccc ccc}
\toprule
    & \multicolumn{3}{c}{Overall IOU} 
    & \multicolumn{3}{c}{Part IOU}
    & \multicolumn{3}{c}{Part transfer (PCP)} \\
\cmidrule(lr){2-4}\cmidrule(lr){5-7}\cmidrule(lr){8-10}
    Method & Horse & Cow & Sheep & Horse & Cow & Sheep & Horse & Cow & Sheep
    \\
\midrule
    SCOPS~\cite{hung2019scops} & 62.9 & 67.7 & 63.2 & 23.0 & 19.1 & 26.8 & - & - & -
    \\
    DINO clustering~\cite{amir2021deep} & 81.3 & 85.1 & 83.9 & 26.3 & 21.8 & 30.8 & - & - & -
    \\
    \midrule
    3D Safari~\cite{zuffi2019three} & 72.2 & 71.3 & 70.8 & - & - & - & 71.7 & 69.0 & 69.3
    \\
    A-CSM~\cite{kulkarni2020articulation} & 72.5 & 73.4 & 71.9 & - & - & - & 73.8 & 71.1 & 72.5
    \\
    LASSIE & 81.9 & 87.1 & 85.5 & 38.2 & 35.1 & \bf43.7 & 78.5 & 77.0 & 74.3
    \\
    \midrule
    Hi-LASSIE w/o inst. MLPs & 80.4 & 80.0 & 79.5 & 30.2 & 29.3 & 33.8 & 76.4 & 74.9 & 71.1
    \\
    Hi-LASSIE w/o feat. MLPs & 82.5 & 87.6 & 85.9 & 34.9 & 32.4 & 39.7 & 74.7 & 72.4 & 74.9
    \\
    Hi-LASSIE w/o $\mathcal{L}_{part}$ & 81.6 & 84.7 & 83.8 & 38.6 & 35.2 & 43.6 & 79.2 & 77.4 & 72.0
    \\
    Hi-LASSIE & \bf83.4 & \bf88.1 & \bf86.3 & \bf39.0 & \bf35.3 & 43.4 & \bf79.9 & \bf77.8 & \bf75.5
    \\
\bottomrule
\end{tabular}
\label{tab:quantitative}
% \vspace{-3mm}
\end{table*}
%%%%%%%%%%%%%%%%%%%%%%%%%%%%%%%%%%%%%%%%%%%%%%%%%%%%%%%%%%%%

\vspace{1mm}
{\noindent \textbf{Keypoint transfer.}~}
Without ground-truth 3D annotations in our datasets,
we follow a common practice~\cite{zuffi2019three,kulkarni2020articulation} to evaluate 3D reconstruction by transferring 2D keypoints from source to target images.
That is, we map a set of 2D keypoints on a source image onto the canonical 3D part surfaces, and project them to a target image via the estimated camera, pose, and shape.
Since the keypoints are transferred from 2D-to-3D and from 3D-to-2D, a successful transfer indicates accurate 3D reconstruction on both the source and target images.
In Table~\ref{tab:keypoint}, we report the percentage of correct keypoints (PCK) under a tight threshold 0.05$\times$max$(h,w)$, where $h$ and $w$ are image height and width, respectively.
The results show that Hi-LASSIE achieves higher PCK on most animal image ensembles compared to the baselines while requiring minimal user inputs.
We also show ablative results of Hi-LASSIE without the instance part MLPs (frequency decomposition), feature MLPs, or zoomed-in part silhouette loss $\mathcal{L}_{part}$.
All the proposed modules can effectively increase the accuracy of keypoint transfer demonstrating their use.

%%%%%%%%%%%%%%%%%%%%%%%%%%%%%%%%%%%%%%%%%%%%%%%%%%%%%%%%%%%%
\vspace{1mm}
{\noindent \textbf{2D overall/part IOU.}~}
In addition to keypoint transfer accuracy, we compare Hi-LASSIE with the baselines using different segmentation metrics (Overall/Part IOU) in Table~\ref{tab:quantitative}.
% other quantitative metrics like 2D overall IOU and part IOU in Table~\ref{tab:quantitative}
%
For Hi-LASSIE and prior 2D co-part segmentation methods like SCOPS~\cite{hung2019scops} and DINO clustering~\cite{amir2021deep}, we manually assign each discovered part to the best matched part in the Pascal-part annotations.
%
% Similar to LASSIE,
Hi-LASSIE can produce accurate overall and part masks in 2D by learning high-fidelity 3D shapes.
Compared to prior methods, Hi-LASSIE outputs match the Pascal-part segmentation better and achieves consistently higher overall IOU.

%%%%%%%%%%%%%%%%%%%%%%%%%%%%%%%%%%%%%%%%%%%%%%%%%%%%%%%%%%%%
\vspace{1mm}
{\noindent \textbf{Part transfer.}~}
Finally, we show the results of part transfer accuracy in Table~\ref{tab:quantitative} using the percentage of correct pixels (PCP) metric proposed in LASSIE~\cite{yao2022lassie}.
The PCP metric is designed similarly as PCK for keypoint transfer, but it uses 2D part segmentations to more densely evaluate 3D reconstruction.
In short, we densely transfer the part segmentation from source to target images through mapping 2D pixels and 3D canonical surfaces.
A correct transfer is done when a pixel is mapped to the same 2D part in both the source and target images.
The PCP results further demonstrate the favorable performance of Hi-LASSIE against prior arts.

%%%%%%%%%%%%%%%%%%%%%%%%%%%%%%%%%%%%%%%%%%%%%%%%%%%%%%%%%%%%
\vspace{1mm}
{\noindent \textbf{Applications.}~}
Hi-LASSIE not only produces high-fidelity 3D shapes but also enables various part-based applications due to explicit skeleton and part-based representation.
For instance, we can easily transfer/interpolate the 3D skeleton transformations to repose or animate the output 3D shapes.
Likewise, we can transfer the surface texture or part deformation from one animal species/instance to another.
Due to their explicit nature, Hi-LASSIE 3D shapes can also be used by graphics artists for downstream applications in AR/VR/games.
We show some example results of these applications in the supplemental material.

\vspace{-2mm}
\section{Conclusion}
\vspace{-2mm}

We propose Hi-LASSIE, a technique for 3D articulated shape reconstruction from sparse image ensemble without using any 2D/3D annotations or templates.
% in this paper to extend the LASSIE optimization framework for higher-fidelity 3D shapes of each animal body.
%
% Unlike prior arts which assume given 3D shapes or skeleton, 
Hi-LASSIE automatically discovers 3D skeleton based on a single reference image from the input ensemble.
We further design several optimization strategies to reconstruct high-resolution and instance-varying details of 3D part shapes across the given ensemble.
Our results on Pascal-Part and LASSIE image ensembles demonstrate the favorable reconstructions of Hi-LASSIE against prior arts despite using minimal user annotations.
In future, we hope to apply Hi-LASSIE on more general articulated objects in-the-wild.

\clearpage
%%%%%%%%%%%%%%%%%%%%%%%%%%%%%%%%%%%%%%%%%%%%%%%%%%%%%%%%%%%%
{\small
\bibliographystyle{ieee_fullname}
\bibliography{egbib}
}

\end{document}